\documentclass[twoside,leqno,twocolumn]{article}

\usepackage[letterpaper]{geometry}
\usepackage{setspace}
\usepackage{ltexpprt}

\usepackage{algorithmicx}
\usepackage{listings}
\usepackage{hyperref}

\usepackage{caption}
\usepackage{subcaption}
\usepackage{siunitx}
\usepackage{etoolbox}
\robustify\bfseries

\usepackage[square,numbers,sort]{natbib}
\usepackage{booktabs}

\usepackage{amsmath}
\usepackage{amssymb}
\usepackage{graphicx}
\usepackage{xspace}

\newcommand{\defemph}[1]{\textbf{#1}}

\DeclareMathOperator*{\argmax}{arg\,max}
\newcommand{\func}[3]{#1\! : #2 \to #3}
\newcommand{\card}[1]{\left|#1\right|}
\newcommand{\abs}[1]{\left|#1\right|}
\newcommand{\derivat}[3]{\frac{\mathrm{d}#1}{\mathrm{d}#2}\Bigr|_{\substack{#2=#3}}}

\newcommand{\R}{\mathbb{R}}
\newcommand{\inspace}{\mathbb{X}}
\newcommand{\outspace}{\mathbb{Y}}

\newcommand{\prob}[1]{P(#1)}
\newcommand{\condprob}[2]{P(#1 \mid #2)}
\newcommand{\expect}[1]{E(#1)}
\newcommand{\condexp}[2]{E(#1 \mid #2)}

\newcommand{\queries}{\mathcal Q}
\newcommand{\props}{\mathcal P}
\newcommand{\indic}{\mathbf 1}

\newcommand{\risk}{L}
\newcommand{\emprisk}{\hat{L}}
\newcommand{\empregrisk}{\hat{L}_\lambda}
\newcommand{\extent}[1]{I(#1)}
\newcommand{\obj}{\mathrm{obj}}
\newcommand{\oest}{\mathrm{bnd}}

\newcommand{\empimp}[2]{#1 \rightarrow #2}
\newcommand{\notempimp}[2]{#1 \not\rightarrow #2}
\newcommand{\empequ}[2]{#1 \leftrightarrow #2}

\newcommand{\crit}{\mathrm{crt}}
\newcommand{\tail}[1]{\mathrm{tail}(#1)}
\newcommand{\prefix}{\sqsubseteq}

\newcommand{\dataset}[1]{\textit{#1}}
\newcommand{\rulefit}{\textit{RuleFit}\xspace}
\newcommand{\xgboost}{\textit{XGBoost}\xspace}

\begin{document}

\title{\Large Better Short than Greedy:\\
Interpretable Models through Optimal Rule Boosting}
\author{Mario Boley\thanks{Monash University, mario.boley@monash.edu}
\and Simon Teshuva\thanks{Monash University, simon.teshuva@monash.edu}
\and Pierre Le Bodic\thanks{Monash University, pierre.lebodic@monash.edu}
\and Geoffrey I. Webb\thanks{Monash University, geoff.webb@monash.edu}
}

\date{}

\maketitle

\fancyfoot[R]{\scriptsize{Copyright \textcopyright\ 2021 by SIAM\\
Unauthorized reproduction of this article is prohibited}}

\begin{abstract} \small\baselineskip=9pt
Rule ensembles are designed to provide a useful trade-off between predictive accuracy and model interpretability. However, the myopic and random search components of current rule ensemble methods can compromise this goal: they often need more rules than necessary to reach a certain accuracy level or can even outright fail to accurately model a distribution that can actually be described well with a few rules. Here, we present a novel approach aiming to fit rule ensembles of maximal predictive power for a given ensemble size (and thus model comprehensibility). In particular, we present an efficient branch-and-bound algorithm that optimally solves the per-rule objective function of the popular second-order gradient boosting framework. Our main insight is that the boosting objective can be tightly bounded in linear time of the number of covered data points. Along with an additional novel pruning technique related to rule redundancy, this leads to a computationally feasible approach for boosting optimal rules that, as we demonstrate on a wide range of common benchmark problems, consistently outperforms the predictive performance of boosting greedy rules.
\end{abstract}

\section{Introduction}
\begin{figure*}[t]
    \centering
    \begin{subfigure}[c]{0.32\textwidth}
    \centering
    \includegraphics[width=\textwidth]{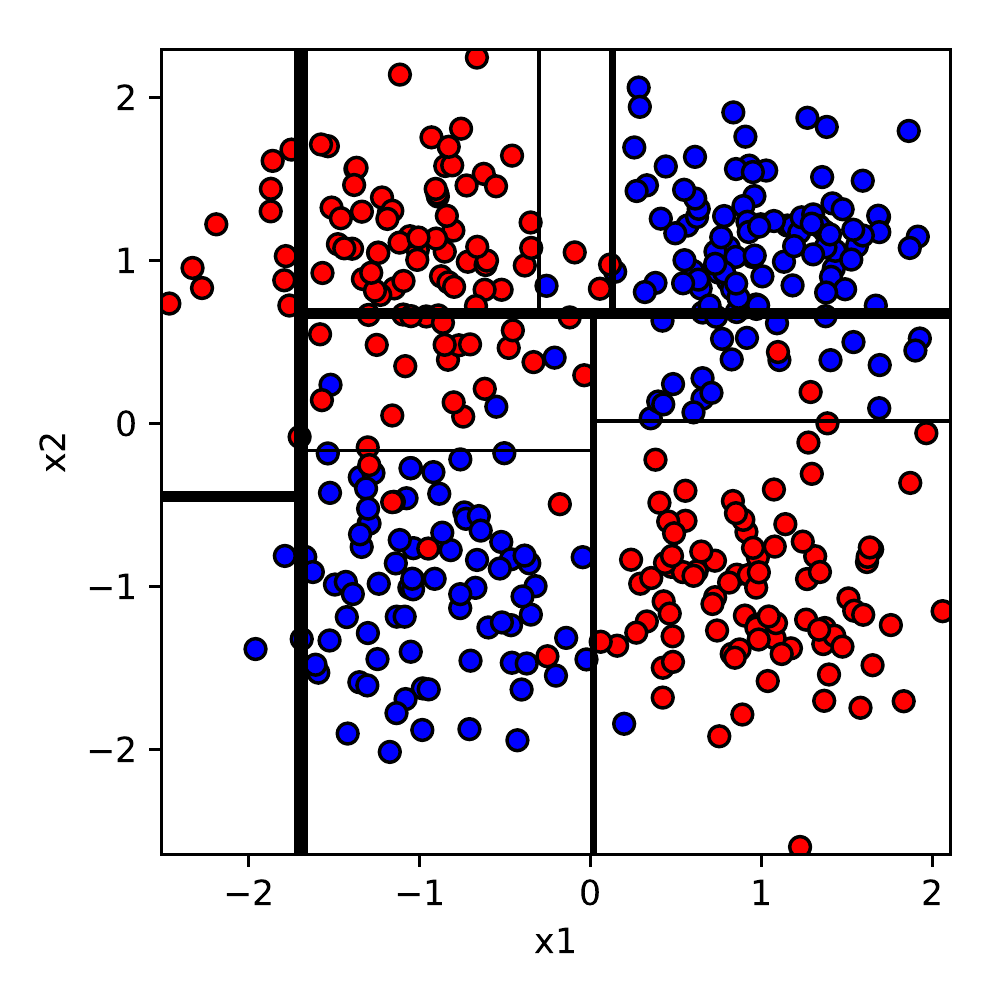}
    \caption{greedy tree cuts, $d=2$, $n=400$}
    \label{subfig:2d_sample}
    \end{subfigure}
    \hfill
    \begin{subfigure}[c]{0.32\textwidth}
    \centering
    \includegraphics[width=\textwidth]{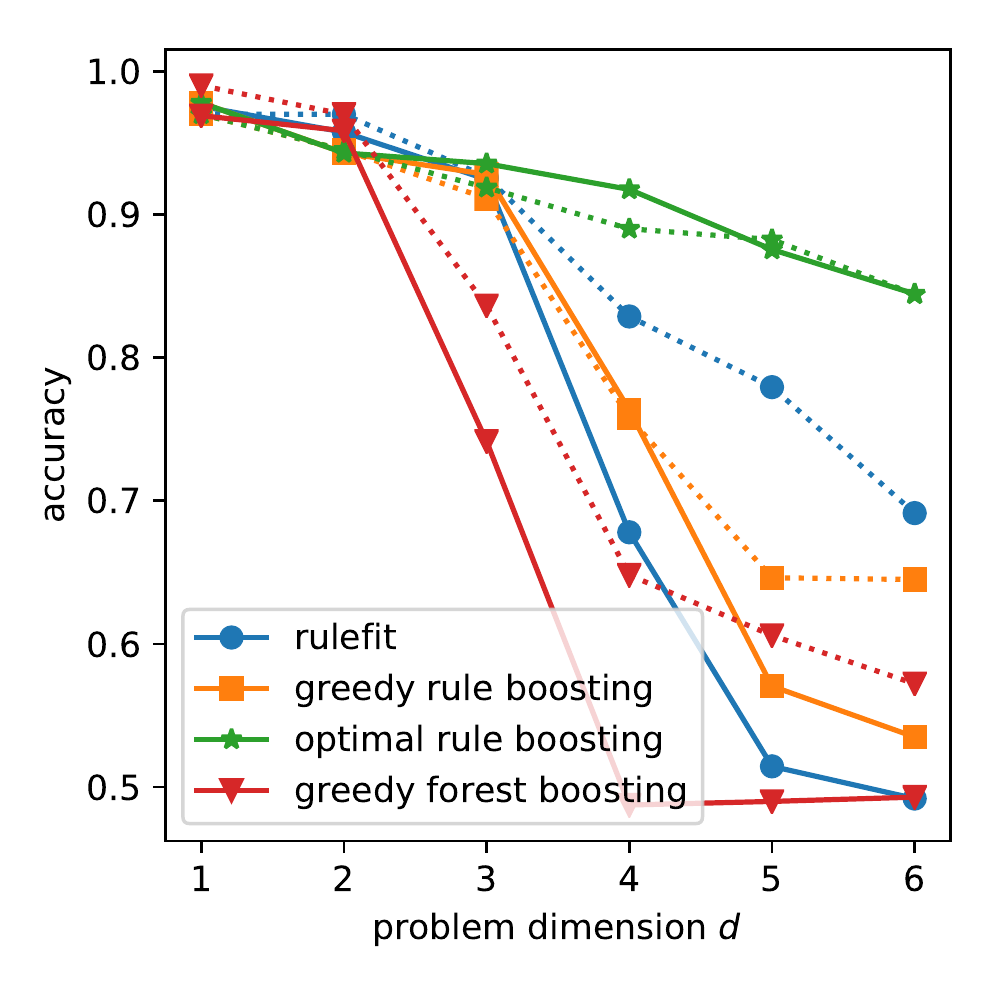}
    \caption{acc. for var. $d$, $n=100 \! \times\! 2^d$, $k=2^d$}
    \label{subfig:dim_vs_acc}
    \end{subfigure}
    \hfill
    \begin{subfigure}[c]{0.32\textwidth}
    \centering
    \includegraphics[width=\textwidth]{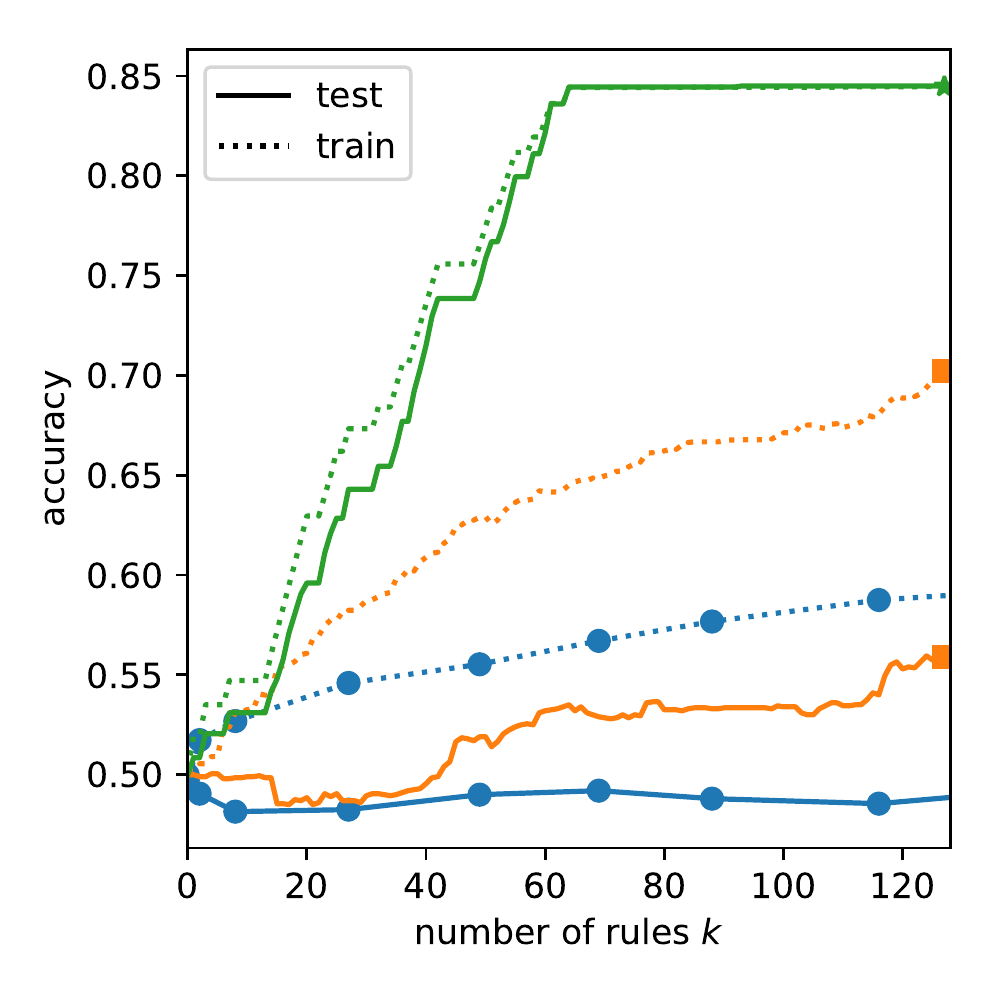}
    \caption{acc. for var. $k$, $d=6$, $n=6400$}
    \label{subfig:num_rules_vs_acc}
    \end{subfigure}
    \caption{Weakness of heuristic rules in \textit{noisy parity problem} defined by latent cluster centers
    $C \sim \mathrm{Unif}(\{-1, 1\}^d)$, observed features $X \mid C \sim \mathrm{Norm}(C, \sigma^2 I_d)$, and target variable $Y \mid C = \prod_{i=1}^d C_i$. Greedy cuts lead to sub-optimal partitioning already for $d=2$ (\subref{subfig:2d_sample}).
    Since there is no association between the target variable and any proper subset of features variables, i.e., $\condprob{Y}{X_{i_1},\dots,X_{i_l}} = \prob{Y}$ for all $\{X_{i_1},\dots,X_{i_l}\} \subset \{X_1, \dots, X_d\}$, the first $d-1$ greedy choices are dictated by sample properties that are non-representative for the underlying distribution. 
    In contrast, gradient boosting ensembles with optimal rules approximate with high probability
    the 0/1-risk minimizing $2^d$ rules
    ``$(-1)^{\card{N}} \, \text{if} \, \bigwedge_{i \in N} X_i \leq 0 \bigwedge_{i \in P} X_i > 0$'' for all $N \subset \{1, \dots, d\}$ and $P = \{1, \dots, d\} \setminus N$.
    Correspondingly, optimal rules retain non-trivial accuracy for growing $d$ whereas the performance of heuristic rules quickly deteriorates to the level of random guessing (\subref{subfig:dim_vs_acc}).
    Importantly, the performance of the heuristic ensembles is not improved for $n \to \infty$ but only with a substantial increase of the number of ensemble members $k$ (\subref{subfig:num_rules_vs_acc}).
    }
    \label{fig:parity_example}
\end{figure*}

Rule learners are designed to deliver models that are interpretable and at the same time have a predictive performance that is competitive with complex tree ensembles.
In particular, the gradient boosting approach provides a theoretically well-founded framework to combine simple prediction rules into powerful additive ensembles.
However, the greedy and random search techniques that
are traditionally employed to fit ensemble members can compromise model comprehensibility or, even worse, outright fail to adequately learn a distribution that can be described well with relatively few rules.
This is because heuristically found rules tend to not fully capture higher order feature interactions, at least not in the simplest way possible (see Fig.~\ref{fig:parity_example}). Consequently, a greater number of rules is required to achieve a certain predictive performance.

As a remedy, this paper presents an optimal rule fitting procedure as base learner for gradient boosting
and, 
contrary to the traditionally raised concerns, 
shows that:
(i) \textit{finding optimal rules in a boosting iteration is computationally feasible}---at least on common benchmark problems when targeting comprehensible ensembles with a small number of rules, and
(ii) \textit{optimal rules consistently outperform greedily found rules in terms of their predictive performance}---an effect that is again more pronounced for smaller ensemble sizes.
Thus, the proposed algorithm is an important extension to the gradient boosting toolbox, in particular for fitting small interpretable rule ensembles.

\subsection{Related Work}
Most predictive rule learning systems are \textit{stagewise fitting} algorithms that identify individual rules one at a time based on their ability to improve the predictive performance of previously fixed rules.
To achieve this, early \textit{sequential coverage} (or \textit{separate-and-conquer}) algorithms~\cite[see survey][]{furnkranz1999separate} simply remove data points covered by already selected rules before fitting subsequent stages, which results in rule lists that represent nested IF-ELSEIF-\dots-ELSE structures.
These structures are typically hard to interpret because the effect of an individual rule has to be understood in the context of the preceding rules.

\textit{Additive rule ensemble} algorithms instead produce flat sets of IF-THEN-rules, the outputs of which are simply added up to find the prediction for a given data point. Thus, individual rule effects on a prediction can be assessed in isolation.
Examples for this approach are recent methods based on greedy sub-modular optimization~\citep{lakkaraju2016interpretable, zhang2020diverse}, earlier local pattern based methods~\citep[see][]{furnkranz2010guest}, as well as \rulefit~\citep{friedman2008predictive}.
These methods build additive ensembles by performing a sub-selection from a set of candidate rules that are obtained in an initial pre-processing step.
\rulefit is a particularly popular approach that overcomes the stagewise fitting paradigm and instead globally optimizes a sparse coefficient vector for the candidate rules.
However, a general downside of fixing a set of pre-processed candidate rules is that this set might not contain optimal building blocks (e.g., the \rulefit candidates are based on greedy tree ensembles).

An alternative line of methods~\citep{cohen1999simple, dembczynski2008maximum, dembczynski2010ender} produces additive rule ensembles by adopting the \textit{gradient boosting} framework~\citep{friedman2001greedy, chen2016xgboost}. This is again a stagewise process, in which new rules are fitted based on their effect on the training loss when their output is added to the prediction scores of the previously fixed rules.
This approach is universally applicable to various kinds of learning problems---provided that the predictive performance for an individual training point can be quantified by a differentiable loss function.
Moreover, it does not require a bounded candidate set and instead optimizes in each stage a  statistically sound objective function over all available conjunctive queries.

Notably though, current rule boosting methods inherited from their sequential coverage ancestors the heuristic search methods for fitting individual rules, i.e., greedy, beam, and stochastic search.
Originally, this preference was not only motivated by computational efficiency but also to avoid overfitting (through ``over-searching'' \cite[see][Sec.~4.1.3]{furnkranz1999separate}).
Given that boosting solves this issue in a more principled way, the exact optimization of the rule objective now appears unambiguously desirable---if the computational concern can be addressed.

While the statistical learning literature tends to avoid solving hard discrete optimization problems, the knowledge discovery and data mining literature often suggests to solve hard rule discovery problems exactly by exhaustive branch-and-bound search.
Here, the conjunctive rule lattice is searched by iteratively augmenting conjunctions with further conditions/propositions (branch), but augmentations are skipped when they cannot lead to an improvement over the currently discovered optimum (bound).
While this approach has a non-polynomial worst-case time complexity, there are a number of techniques that make it practically applicable to many datasets.

One idea, introduced in the \textit{OPUS (Optimized Pruning for Unordered Search) framework}~\citep{webb1995opus, webb2001discovering}, is to only check an augmentation if it successfully passed the pruning check on the parent level of the current search node.
This technique leverages the monotonicity of the bounding function and usually achieves speed-ups proportional to the number of available propositions.

A further development is the introduction of \emph{tight bounding functions} (also ``tight optimistic estimators'') \citep{morishita2000transversing, grosskreutz2008tight}
that bound the value of all possible refinements of a conjunction by identifying the optimal sub-selection of the data points selected by that conjunction. While this approach disregards whether this sub-selection can be described by an actual refinement, it is much more effective than simple term-wise upper bounding (that one obtains by more straightforward adaptions of support-based pruning) and, compared to those, reliably leads to orders of magnitude speed-up~\cite{lemmerich2016fast}.

A final crucial technique is to eliminate redundancies in the explored part of the rule language by searching among \emph{rule equivalence classes} where two rules are considered equivalent if they select the same data.
The roots of this notion are in the formal concept analysis and knowledge discovery literature~\citep{stumme2002computing, bastide2000mining} as a ``condensed representation'' of the information contained in a data collection.
However, when used in conjunction with efficient traversal techniques~\citep{uno2004efficient}, considering equivalence classes of rules, naturally also leads to substantial speed-ups in rule optimization similar to the gains from tight bounding functions~\cite{boley2009non, boley2017identifying}.



\subsection{Contributions}
This paper builts on the above-mentioned optimization techniques to devise an efficient optimal base learner for gradient boosting. The resulting rule learner outperforms greedy rule boosting as well as \rulefit in terms of the number of rules required for reaching a specific predictive performance. 
Thus, the presented approach is ideally suited to produce comprehensible rule ensembles.
In detail:
\begin{enumerate}
\item We derive the per-stage rule objective function for second order gradient boosting 
and, as main theoretical result, show that the objective values of refinements of a selected query can be tightly upper-bounded in linear time in the number of selected data points (Thm~\ref{thm:greedy_bnd}, Sec.~\ref{subsec:pruning}).
\item We integrate this bound into an anytime base learner for rule boosting based on branch-and-bound (Sec.~\ref{subsec:condensation}). 
As innovation of independent interest, we give a formulation of rule equivalence class search that can be exploited in the OPUS propagation of pruning information (Thm.~\ref{thm:pp_propagation}). 
\item We empirically evaluate the resulting optimal rule boosting algorithm by comparing its predictive performance as well as its computational efficiency across 24 common benchmark classification and regression problems (Sec.~\ref{sec:evaluation}).
\end{enumerate}
We find that the optimal rule learner consistently outperforms the predictive performance of conventional greedy rule boosting as well as \rulefit when targeting comprehensible ensembles of $10$ rules or less.
Moreover, while its computational demands are naturally higher than that of myopic search, it does not require more than 2h (and is usually considerably faster) across all considered benchmarks---even using a standard PC and a preliminary Python implementation (\url{https://github.com/marioboley/realkd.py}).

\section{Gradient Rule Boosting}

\subsection{Predictive Modeling}
Formally, we aim to model a \defemph{target variable} $Y \in \outspace$ given some \defemph{feature vector} $X \in \inspace$ based on \defemph{training data} $\{(x_i, y_i)\}_{i=1}^n$ that has been sampled according to the joint distribution of $X$ and $Y$.
We focus on models in the form of a single-valued scoring function $\func{f}{\inspace}{\R}$.
For instance, in regression problems ($\outspace=\R$),  $f$ typically models the conditional expectation of the target, i.e.,  $f(x) \approx \condexp{Y}{X=x}$, whereas in binary classification problems ($\outspace=\{-1,1\}$), $f$ typically models the conditional \defemph{log odds}, i.e., $f(x) \approx \ln \condprob{Y=1}{X=x}/\condprob{Y=-1}{X=x}$ and the conditional probabilities $p(y \mid x)$ are recovered by the sigmoid transform
\[
p(y \mid x) = \sigma(f(x))=(1 + \exp(-yf(x)))^{-1} \enspace .
\]

Generally, the meaning of a score $f(x)$ is encapsulated in a positive \defemph{loss function}
$l(y, f(x))$ that represents the cost of predicting $f(x)$ when the true target value is $y$. Specific examples are the \defemph{squared loss} $l(y,f(x))=(y-f(x))^2$ for regression problems and the \defemph{logistic loss} $l(y,f(x))=\log(1 + \exp(-yf(x)))$ for classification problems.
However, we only assume that $l$ is twice differentiable and convex in its second argument.
In general, our goal is to find a model $f$ that minimizes the prediction \defemph{risk}, i.e., the expected loss according to the underlying distribution: $\risk(f) = \expect{l(y, f(x))}$.
Since we have no information about that distribution other than the given training data, a learning algorithm can only compute the \defemph{empirical risk}:
\begin{equation*}
    \emprisk(f) = \frac{1}{n} \sum_{i=1}^n l(y_i,f(x_i)) \enspace .
\end{equation*}
However, naively optimizing the empirical risk might result in overfitting, i.e., a model with a poor actual prediction risk.
To avoid this, learning algorithms typically optimize the \defemph{regularized (empirical) risk}
\begin{equation}
\empregrisk(f) = \emprisk(f) + \frac{\lambda}{n} \Omega(f)
\label{eq:empregrisk}
\end{equation}
where the empirical risk term is balanced with a model complexity term $\Omega$ weighted by a non-negative \defemph{regularization parameter} $\lambda \geq 0$. As we will see below, a  positive regularization parameter $\lambda > 0$ is not only useful from a statistical perspective but also required by the gradient boosting framework when working with non-strictly convex loss functions (for which the second derivative can vanish).

\subsection{Additive Rule Ensembles}
The basic functions considered in this paper are \defemph{rules}
$r(x)=wq(x)$ where $q: \inspace \to \{0, 1\}$ is a binary \defemph{query} function and $w \in \mathbb{R}$ is a \defemph{prediction weight}, which correspond to the rule's antecedent and consequent, respectively. 
The \defemph{extent} (coverage) of a rule/query are the indices of the instances selected by the query, i.e., $\extent{q}=\{i: q(x_i)=1, 1\leq i \leq n\}$.
These functions represent IF-rules that predict the numeric value $w$ for an input $x$ if the condition represented by $q$ holds for $x$ and $0$ otherwise (again, the interpretation of the predicted value depends on the modeling task). 

Specifically, as possible antecedents, we consider the class of \defemph{conjunctive queries} that can be formed from a set of ``atomic'' \defemph{propositions} $\props \subseteq \{0,1\}^\inspace$, i.e., 
\begin{equation*}
\queries = \left\{q \in \{0,1\}^\inspace: q(x) = \prod_{p \in \props(q)} p(x), \enspace \props(q) \subseteq \props \right\}
\end{equation*}
where $\props(q)$ denotes the subset of propositions contained in query $q$.
For instance, for the common case of $\inspace = \R^d$ one usually chooses threshold functions 
\begin{align*}
\props=\{ &
\indic(x^{(j)} \leq x^{(j)}_i),
 \indic(x^{(j)} > x^{(j)}_i):\\
 &1 \leq j \leq d, 1 \leq i \leq n \}
\end{align*}
as propositions.
With this choice, $\queries$ corresponds to the set of convex polytopes with axis-parallel sides. However, more general constructions are possible. In particular, it is easy to accommodate input spaces that mix continuous with categorical dimensions.  

By forming sums of individual rules, we can then form the function class of \defemph{additive rule ensembles}
\begin{equation*}
    \mathcal{F}= \left\{f \in \R^{\inspace}: f(x)=\sum_{i=1}^k w_i q_i(x), \enspace q_i \in \queries, w_i \in \R \right\}
\end{equation*}
where $k$ is the \defemph{ensemble size}.
Note that, assuming that $\props$ is closed under negation,
tree models and by extension additive tree ensembles are defined using the same function class.
However, each individual tree model has the restriction to be a sum of rules where the queries partition $\inspace$. This typically results in a much larger number of rules to accurately describe a given distribution \citep{fan2018rectangular}.

\rulefit finds a function in $\mathcal{F}$ by restricting $\queries$ to a candidate query set of manageable size $k'$ and then minimizing~\eqref{eq:empregrisk} over $\R^{k'}$ via convex optimization.
To facilitate a small effective ensemble size $k$, the sparsity-inducing l1 complexity-measure $\Omega(f)=\sum_{i = 1}^k \abs{w_i}$ is used.
In contrast, \defemph{rule boosting} grows an additive ensemble of arbitrary desired size by starting from the empty model $f_0 \equiv 0$ and then iteratively calling a \defemph{base learner} to find a new term $w_t q_t$ that minimizes the regularized loss when combined with the previously fixed part of the model $f_{t-1}=\sum_{i=1}^{t-1} w_i q_i$.


\subsection{Objective Function}
Here, we adapt the ``second order'' gradient boosting approach popularized by \xgboost~\cite{chen2016xgboost}:
We define the model complexity as the squared Euclidean norm of the rules' consequents: $\Omega(f)=\sum_{i = 1}^k w^2_i$. 
Then, to estimate the effect of $r_t$ on \eqref{eq:empregrisk} in a computationally convenient way,
we approximate the regularized risk of $f_t$ as
\begin{align*}
\empregrisk(f_t) &= \frac{\lambda}{2n} \sum_{i=0}^t w_i^2 + \frac{1}{n} \sum_{i=1}^n l(y_i, f_{t-1}(x_i) + w_t q_t(x_i))\\
&\simeq \empregrisk(f_{t-1}) + \frac{w_t}{n}\! \left(\sum_{i \in I(q_t)}\! g_i \right)\! + \frac{w^2_t}{2n}\!\left(\!\lambda +\hspace{-0.3cm} \sum_{i \in I(q_t)} \!h_i\!\right) 
\end{align*}
where we approximated the loss $l(y_i, f_{t-1}(x_i) + w_t q_t(x_i))$ by $l(y_i, f_{t-1}(x_i)) + g_i w_t q_t(x_i) + \frac{1}{2}h_i (w_t q_t(x_i))^2$ using the first and second order \defemph{gradient statistics}, $g_i$ and $h_i$, of the prediction loss incurred at example $i$:
\begin{equation*}
    g_i=\derivat{l(y_i, y)}{y}{f_{t-1}(x_i)} \enspace ,
    \quad
    h_i=\frac{\mathrm{d}^2 l(y_i, y)}{\mathrm{d}y^2}\Bigr|_{\substack{y=f_{t-1}(x_i)}}
    \enspace .
\end{equation*}
The objective in the $t$-th iteration of gradient boosting is to minimize the regularized loss of $f_t$, or equivalently to maximize the \defemph{loss reduction}, $ \empregrisk(f_{t-1})-\empregrisk(f_t)$. 
For a fixed rule antecedent $q_t$, the associated optimal weight $w_t$ is given by
\begin{equation}
    w_t = -\frac{\sum_{i \in I(q_t)} g_i}{\lambda + \sum_{i \in I(q_t)} h_i}
    \label{eq:opt_weight}
\end{equation}
Thus, we end up with the following objective function for $q_t$ that we seek to maximize over the possible antecedents $q \in \queries$:
\begin{equation}
    \obj(q) = \frac{\left( \sum_{i \in I(q)} g_i \right )^2}{2n \left(\lambda + \sum_{i \in I(q)} h_i \right)}
    \label{eq:obj}
\end{equation}

For instance, for regression problems using the squared loss we end up with the following gradient statistics and optimal rule weight:
\begin{align*}
    g_i &= -2(y_i - f_{t-1}(x_i)) \\
    h_i &= 2 \\
    w_t &=\sum_{i \in I(q_t)} \frac{\left ( y_i - f_{t-1}(x_i) \right )}{\lambda/2 + \card{I(q_t)}} \enspace ,
    \intertext{and for classification with the logistic loss we have:}
    g_i &= -y_i \, p(-y_i \mid x_i) \\
    h_i &= p(y_i \mid x_i)\, p(-y_i \mid x_i)\\
    w_t&=\frac{\sum_{i \in I(q_t)} y_i \, p(-y_i \mid x_i)}{\lambda + \sum_{i \in I(q_t)} p(y_i \mid x_i) \, p(-y_i \mid x_i)} \enspace .
\end{align*}
For the latter, we can see that only for strictly positive $\lambda$, the optimal rule weight $w_t \rightarrow 0$ when $\prod_{i \in I(q_t)} p(y \mid x_i) \rightarrow 1$, i.e., when for all $i \in I(q_t)$ the modeled probabilities $p(y_i \mid x_i) \rightarrow 1$. 
In contrast, for $\lambda=0$ we have $w_t \rightarrow 1$ even as the above conditional likelihood approaches $1$. Thus, regularization seems very appropriate for this model (though even for $\lambda=0$, $\obj(q_t) \rightarrow 0$ if the conditional likelihood approaches $1$).

\section{An Efficient Optimal Base Learner}

In this section, we develop a practically efficient base learner for rule boosting that maximizes the objective function \eqref{eq:obj} exactly.
As a convention, we fix some arbitrary order of the set of basic propositions $\props = \{p_1, \dots, p_d\}$ and identify queries with ordered subsets of $\props$, i.e., $q=\{p_{i_1}, \dots, p_{i_l}\}$ with $i_{j} < i_{j+1}$ for $1 \leq j < l$. 
The maximum index $i$ such that $p_i \in q$ is called the \defemph{tail index}, denoted $\tail{q}$.
In this framework, we say that a query $q'$ is a \defemph{tail augmentation} of $q$, denoted as $qp_i$, if $q'=q \cup \{p_a\}$ for some $a>\tail{q}$.
Finally, we call a query $q$ a \defemph{prefix} of another query $q'$, denoted by $q \prefix q'$, if $q'$ can be generated from $q$ via successive tail augmentations.

In a nutshell, the proposed base learner maximizes $\obj$ by enumerating candidate queries recursively via tail augmentations starting from the trivial query $q=\top$.
Of course, naively this would always require a number of objective function evaluations exponentially in the number of propositions.
We combat this exponential growth with two techniques: (i) search space pruning by bounding the objective value that can be attained by augmentations of a specific query and (ii) search space condensation by replacing the naive search space $\queries$ with a typically much smaller search space $\queries' \subseteq \queries$ in which the optimal objective value is still attained.

\subsection{Search Space Pruning}
\label{subsec:pruning}
A general construction for effective bounding functions is often referred to as \defemph{tight optimistic estimator} in the rule discovery literature. Here, the term ``tight'' refers to the ``selection-unaware'' scenario where we relax the possible inputs to $\obj$ to include all possible index subsets; instead of just those that can be selected by a query $q$. This leads to the following relaxed function:
\begin{align*}
\oest(q) &= \max\{\obj(J): J \subseteq I(q)\}\\
 &\leq \max\{\obj(q'): q' \supseteq q\} 
 \enspace .
\end{align*}
While this formulation implies that $\oest$ is an admissible pruning function for branch-and-bound search, in general it remains unclear how to compute it efficiently.

Fortunately, for our objective function~\eqref{eq:obj} there is a general algorithm that computes $\oest(I)$ in time $O(\card{I})$ after an initial pre-sorting step of cost $O(n \log n)$ that has to be applied only once for each rule in the ensemble. This algorithm finds an optimal index subset $J^* \subseteq I$ by greedily selecting indices $i \in I$ in order of their corresponding loss derivative ratio $g_i / h_i$. The correctness of this approach is established with the following result.
\begin{theorem}
Let $i_1, \dots i_m$ be the sequence of indices in the set $I \subseteq \{1, \dots, n\}$ ordered such that
\[
\frac{g_{i_1}}{h_{i_1}} \geq \frac{g_{i_2}}{h_{i_2}} \geq \dots \geq \frac{g_{i_m}}{h_{i_m}} \enspace .
\]
Then there is an index subset $J^*$ with $\oest(I)=\obj(J^*)=\max \{\obj(J): J \subseteq I\}$ that is given as a prefix or a suffix of the above sequence, i.e., for some $l>0$
\begin{align}
J^* &= \{i_1, i_2, \dots, i_l\} \quad \text{ and } \quad g_{i_l} > 0 \quad \text{ or }
\label{eq:optprefix}\\
J^* &= \{i_{m-l+1}, \dots, i_m\} \quad \text{ and } \quad g_{i_{m-l+1}} < 0 
\label{eq:optsuffix}
\enspace .
\end{align}%
\label{thm:greedy_bnd}
\end{theorem}
\vspace{-1.2cm}
\begin{proof}
\newcommand{\transpos}{\mathrm{trans}}
First, we can observe that for any optimal index subset $J^*$ we have $J^* \subseteq I_+$ or $J^* \subseteq I_-$ with $I_+=\{i \in I: g_i > 0\}$ and $I_-=\{i \in I: g_i < 0\}$: Indeed, for any $J \subseteq I$ with non-negative gradient sum $G_J=\sum_{i \in J}g_i \geq 0$ and $j \in J \cap I_-$ we have $G_{J \setminus \{j\}} > G_J$. Moreover, due to the positivity of $h_j$, we have $\sum_{i \in J}h_i=H_{J} > H_{J \setminus \{j\}}$, and thus $\obj(J \setminus \{j\}) > \obj(J)$. The same is true for $J$ with $G_J \leq 0$ and $j \in J \cap I_+$. 
Thus, it is sufficient to show \eqref{eq:optprefix} for the case $J^* \subseteq I_+$ and \eqref{eq:optsuffix} for $J^* \subseteq I_-$.
Moreover, by symmetry of the objective function it suffices to show the first case (as $\{i_{m-l+1},\dots,i_m\} \subseteq I_-$ is optimal wrt $g$ and $h$ iff $\{i_{1},\dots,i_l\}\subseteq I_+$ is optimal wrt $-g$ and $h$).

For an index subset $J \subseteq I_+$, let $\transpos(J)$ be the number of transpositions (or sorting violations) of $J$, i.e., the number of ordered index pairs $i, j$ such that $j \in J$ and $i \in I_+ \setminus J$ but $g_i/h_i \geq g_j/h_j$. We will show that for each $J \subseteq I_+$ with $\transpos(J)>0$ there is an index set $J' \subseteq I_+$ with $\transpos(J') < \transpos(J)$ such that $\obj(J') \geq \obj(J)$. Consequently, there is an optimal index set $J^*$ with $\transpos(J^*)=0$ as required.

Let $\transpos(J)>0$ and $j \in J$ and $i \in I_+\setminus J$ an index pair with $g_i/h_i \geq g_j/h_j$ (or equivalently $g_i h_j \geq g_j h_i$). In the special case that $J = \{j\}$ and $\lambda = 0$, we can directly see that, up to a factor $2n$,
\begin{align*}
   \obj(J \cup \{i\}) - \obj(J) & = \frac{(g_i + g_j)^2}{h_i + h_j} - \frac{g_j^2}{h_j}
   \\
   & = \frac{g_i^2 h_j + g_j(2g_i h_j - g_j h_i)}{(h_i + h_j)h_j} \geq 0 \enspace .
\end{align*}
For the case that $\lambda > 0$ or $\card{J} > 1$, we can apply Lemma~A.1
from the supplementary material, which states that for positive numbers $r, a, c$ and strictly positive numbers $s, b, d$ with $a/b \geq c/d$,
\[
\frac{(r+c)^2}{s+d} \leq \max \left\{\frac{r^2}{s}, \frac{(r + a + c)^2}{s + b + d}\right\} \enspace .
\]
By setting $r = \sum_{t \in J \setminus \{j\}} g_t$, $s = \lambda + \sum_{t \in J \setminus \{j\}} h_t$, $a=g_i$, $b=h_i$, $c=g_j$, and $d=h_j$ it follows that 
\[
\obj(J) \leq \max \left\{ \obj(J \setminus \{j\}), \obj(J \cup \{i\}) \right\} \enspace .
\]
Since $\transpos(J \setminus \{j\}) < \transpos(J)$ as well as $\transpos(J \cup \{i\}) < \transpos(J)$, $J \setminus \{j\}$ or $J \cup \{i\}$ is an index set with strictly fewer transpositions that dominates $J$.
\end{proof}


\lstdefinelanguage{paper}{morekeywords={add, append, if, then, while, for, pop, push, enqueue, is, not, pop, from, with, and, init, to, return, sort}, sensitive=false,
morecomment=[l]{\#}}

\renewcommand{\lstlistingname}{Algorithm}
\lstset{language=paper}
\lstset{numbers=left, numberstyle=\tiny, stepnumber=1, numbersep=5pt,
basicstyle=\small,
mathescape=true,
label=alg:branch-and-bound,
flexiblecolumns=true,
numberblanklines=false,
belowcaptionskip=2pt,
escapechar=\@}
\begin{lstlisting}[float=t, frame=lines, 
caption={\emph{Branch-and-Bound Base Learner}.%
% Along with queries $q$ and extents $I$ to be expanded, search nodes enqueued in boundary also keep track of corresponding promising augmentations along with their tightest estimated bound and critical index ($(a,b,c) \in A$), which are used to filter them in line~\ref{line:for}.
},
xleftmargin=10pt,
xrightmargin=5pt,
captionpos=b]
sort data s.t. $g_{i-1}/h_{i-1} > g_{i}/h_i$, $1 < i \leq n$
init boundary to empty priority queue
$q^*=\top$, $I = \{1, \dots, n\}$
$A=\{(i, \infty, i): 1 \leq i \leq d\}$ @\hfill@# (aug idx, bound, crit idx)
push $(q^*, I, A)$ to boundary
while boundary not empty
    pop $(q, I, A)$  from boundary
    init $A'$ to empty set
    for $(a, b, c) \in A$ if $\notempimp{q}{p_a}, \obj(q^*)<b, \tail{q} \leq c$@\label{line:for}@
        $I_a = I \cap \extent{p_a}$ @\hfill@# via merge retaining order
        $q^*=\argmax \{\obj(qp_a), \obj(q^*)\}$
        $b' = \oest(I_a)$ @\hfill@#  via Thm. @\ref{thm:greedy_bnd}@
        $c' = \begin{cases}c &,\text{ if }c < a \text{ and } \notempimp{q}{p_c}\\
        \crit(qp_a) &,\text{ otherwise}\end{cases}$ @\hfill@# (3.7)
        add $(a, b', c')$ to $A'$
    for $(a, b, c) \in A'$ if $a=c$
        push ($qp_a$, $I_a$, $A'$) to boundary
$q_\text{short} = \text{apx. shortest equivalent to } q^*$
return $r(x) = wq_\text{short}$(x) @\hfill@# $w$ via Eq. @\eqref{eq:opt_weight}@@\label{line:shortest}@
\end{lstlisting}
\subsection{Search Space Condensation}
\label{subsec:condensation}
An important observation about the objective function $\obj(q)$ is that it is a function of the 
query extent $I(q)$.
That is, queries $q$, $q'$ with equal extents $I(q)=I(q')$ have the same objective value and can thus be considered (empirically) \defemph{equivalent}, denoted $\empequ{q}{q'}$, for the purpose of optimization.
Similarly, let us say that a query $q$ (empirically) \defemph{implies} another query $q'$, denoted $\empimp{q}{q'}$, if $I(q') \supseteq I(q)$.
The following recursive construction of \citet{uno2004efficient} allows for an efficient enumeration of a single \defemph{core query} per equivalence class:
The trivial query $q = \top$ is a core query.
Moreover, a tail augmentation $qp_i$ of a core query $q$ is also a core query if $\notempimp{q}{p_i}$ and
\begin{equation}
\text{for all } j < i, \text{ if } \empimp{q'}{p_j} \text{ then } \empimp{q}{p_j} \enspace .
\label{eq:prefix_pres}
\end{equation}
It is straightforward to show that the core queries form a prefix tree rooted in $\top$. This allows to enumerate all core queries in a tree traversal where all children of a given tree node $q$ can be found by checking all tail augmentations $qp_i$ and filter for those that satisfy the prefix preservation condition~\eqref{eq:prefix_pres}.

Importantly, the set of core queries is determined by an arbitrary order imposed on the set of propositions $\props$. Also, while core queries are inclusion-minimal in their equivalence class, there can be shorter queries describing a specific extension.
Thus, to encode a generality bias that is independent of the proposition order, it makes sense to convert an optimized core query to an (approximately) shortest equivalent query in a post-processing step (using a greedy algorithm~\citep[see][]{boley2009non}).

While the core query approach leads to a substantial reduction of the search space, the prefix preservation checks have a notable cost (worst case $\Omega(d\card{I(q)})$). 
The following novel notion us to do that: for core query $q$ and tail extension $p_i$ define the \defemph{critical index} of $qp_i$ as
\[
\crit(qp_i) = \min \{j: \empimp{qp_i}{p_j}, \notempimp{q}{p_j}\} \leq i \enspace .
\]
The critical index of a tail augmentation is naturally determined when checking the prefix-preservation condition~\eqref{eq:prefix_pres}, which holds if and only if $\crit(qp_i) = i$.
Knowing the exact critical index can be used to reduce the number of required checks in the successors of certain sibling nodes as shown in the following theorem.
\begin{theorem}
\label{thm:pp_propagation}
Let $q$ and $q'$ be core queries such that $qp_i \prefix q'$ and $c=\crit(qp_j) < j$ for $j > i$. Then
\begin{align}
    &\text{if }\notempimp{q'}{p_c}\text{ then }q'p_j\text{ is not a core query}
    \label{prop:crit_basic}\\
    &\text{if }c < i\text{ then }q'p_j\text{ is not a core query} \enspace .
    \label{prop:crit_rec}
\end{align}
\end{theorem}
\begin{proof}
For property \eqref{prop:crit_basic} we can observe that $\empimp{q'p_j}{p_c}$ because $I(p_c) \supseteq I(qp_j) \supseteq I(q'p_j)$. Together with the premise that $\notempimp{q'}{p_c}$ it follows that $q'p_c$ does not satisfy the prefix condition~\eqref{eq:prefix_pres}.
Property~\eqref{prop:crit_rec} follows as a special case of \eqref{prop:crit_basic} by showing that $c < i$ implies that $\notempimp{q'}{p_c}$.
Indeed, assuming that $\empimp{q'}{p_c}$ there needs to be a prefix-minimal query $q''$ with $qp_i \neq q'' \prefix q'$ such that $\empimp{q''}{p_c}$. However, then $q''$ is not a core query and neither are its suffixes including $q'$, which contradicts our assumptions.
\end{proof}

Algorithm~\ref{alg:branch-and-bound} shows how Thms.~\ref{thm:greedy_bnd} and \ref{thm:pp_propagation} are integrated into the OPUS/branch-and-bound framework.
As usual, every query $q$ that still needs to be considered for augmentation is kept in a priority queue (boundary) along with a list of augmentation elements ($A$) that are promising for that query. 
As novel extension, this list does not only contain the augmentation elements ($a$) and bounds ($b$) to their objective value but also their previously determined critical index ($c$).
This information is used in line~\ref{line:for} to prune augmentations if they lead to (i) equivalent queries ($\empimp{q}{p_a}$), (ii) to queries of insufficient objective value ($\obj(q^*) \geq b$), or (iii) to non-core queries ($\tail{q}>c$).
The latter condition is correct due to~\eqref{prop:crit_rec}, which allows us to recursively prune all refinements $q' \sqsupseteq qp_a$.
Additionally, property~\eqref{prop:crit_basic} allows to skip the prefix preservation check for $qp_a$ itself, as it is guaranteed to lead to a non-core query, although $a$ can still be a valid augmentation index for extensions of $qp_a$.
Finally, in line~\ref{line:shortest} the result query is converted to a shortest equivalent query.

We close the description of the algorithm with noting an important property that helps coping with hard inputs:
It is an anytime algorithm in the sense that we can obtain the current $q^*$ as an approximation to the optimal query at any time when terminating early.
Importantly, we also obtain the \defemph{multiplicative approximation guarantee} $\obj(q^*)/\max \{b: (a,b,c) \in \text{boundary}\}$ for this current guess.
In fact, when we are a priori satisfied with an $\alpha$-approximation to the optimal query, we can directly relax the corresponding condition in line~\ref{line:for} for even more effective pruning.

\section{Evaluation}
\label{sec:evaluation}
\sisetup{
round-mode = places,
round-precision = 1,
detect-weight=true,detect-inline-weight=math
}
\begin{table*}[t]
\centering
\begin{small}
\begin{tabular}{
l
c
c
S[round-precision = 3]
S[round-precision = 3]
S[round-precision = 3]
S[round-precision = 1]
S[round-precision = 1]
S[round-precision = 1]
S[round-precision = 3]
S[round-precision = 1]}
\toprule
 \bf dataset & \bf feat. & \bf rows  & \multicolumn{3}{c}{\bf \#rules/score AUC} & \multicolumn{3}{c}{\bf comp. time (s)} & \multicolumn{2}{c}{\bf unbnd. rf}\\
& & & {rf} & {grd} & {opt} & {rf} & {grd} & {opt} & {score} & {\#rul.}\\
\midrule
breast cancer & 30 & 569 & 0.9080301162 & 0.9493185592 & \bfseries 0.9525430612 & 2.985281229 & 13.94162149 & 44.88710232 & 0.9841652259 & 46\\
digits (5) & 64 & 3915 & 0.5 & 0.9225268252 & \bfseries 0.9274983531 & 10.01824641 & 266.4437896 & 947.6818465 & 0.9760854065 & 228\\
gender recog & 20 & 3168 & 0.9396237946 & 0.9420448153 & \bfseries 0.9562055046 & 7.904347849 & 76.99409242 & 1642.91957 & 0.9825399135 & 113\\
IBM HR & 32 & 1470 & 0.5 & 0.6773357577 & \bfseries 0.6786543228 & 4.103605652 & 44.06386228 & 5925.012461 & 0.6741250583 & 173\\
iris (1) & 4 & 150 & 0.688751491 & 0.9097114778 & \bfseries 0.9152419141 & 1.329851007 & 2.482771587 & 10.06922712 & 0.9436982565 & 99\\
wine (1) & 13 & 178 & 0.793938137 & 0.9311271372 & \bfseries 0.9417542897 & 1.505932617 & 3.153590775 & 3.989798641 & 0.9778650591 & 85\\
classification2 & 8 & 2000 & 0.7531736409 & 0.8690371657 & \bfseries 0.8871691372 & 5.256789708 & 27.16434093 & 228.1542704 & 0.9062428767 & 155\\
telco churn & 18 & 7043 & 0.5 & 0.7762759049 & \bfseries 0.7789425898 & 16.31104407 & 82.58856215 & 517.2982465 & 0.7251912776 & 88\\
tic-tac-toe & 27 & 958 & 0.6993205664 & 0.7519920356 & \bfseries 0.8035371001 & 3.078974342 & 14.43711905 & 15.60955868 & 0.990194092 & 194\\
titanic & 7 & 1043 & 0.6987474776 & 0.857436385 & \bfseries 0.8594484074 & 2.950261879 & 14.7348032 & 144.0979695 & 0.8372092797 & 35\\
\midrule                                                
boston & 13 & 506 & 0.1625784834 & 0.5443269313 & \bfseries 0.5645053743 & 2.408338833 & 10.0941659 & 585.0828248 & 0.8765963453 & 151\\
demographics & 13 & 6876 & 0.2086817046 & 0.3348054207 & \bfseries 0.3431530096 & 14.51268444 & 112.2525083 & 941.6020743 & 0.529089414 & 170\\
insurance & 6 & 1338 & 0.2246165228 & 0.7470574086 & \bfseries 0.7511195371 & 3.630008828 & 11.97364821 & 16.16471052 & 0.835221621 & 236\\
load diabetes & 10 & 442 & 0.1823464846 & 0.2947112798 & \bfseries 0.307904867 & 2.053992693 & 5.721149635 & 51.90929914 & 0.4555706348 & 53\\
friedman1 & 10 & 2000 & 0.2089793687 & 0.5080347367 & \bfseries 0.5530911483 & 5.875086764 & 24.8551085 & 51.35124912 & 0.9771373199 & 231\\
friedman2 & 4 & 10000 & 0.2654882101 & 0.787816404 & \bfseries 0.8075392713 & 5.778061068 & 16.91041012 & 18.52501512 & 0.9989542153 & 638\\
friedman3 & 4 & 5000 & 0.1613169838 & 0.5079139727 & \bfseries 0.523656834 & 4.781754563 & 16.93031182 & 22.49199314 & 0.8812569524 & 192\\
mobile prices & 20 & 2000 & 0.1317269736 & 0.770110641 & \bfseries 0.778511755 & 7.168856151 & 39.49707899 & 2437.309162 & 0.9338768035 & 105\\
red wine quality & 11 & 1599 & 0.1089811642 & 0.2559895966 & \bfseries 0.2673204677 & 4.137505213 & 22.00216656 & 623.1062929 & 0.4485413625 & 138\\
suicide rates & 5 & 27820 & 0.2061492711 & 0.2980792778 & \bfseries 0.3016160648 & 45.2805979 & 243.0924582 & 259.8381851 & 0.39642523 & 249\\
used cars & 4 & 1770 & 0.2690953086 & 0.688876491 & \bfseries 0.7204209694 & 4.195012184 & 12.48066509 & 13.30435801 & 0.9622936196 & 283\\
videogamesales & 6 & 16327 & 0.1160189281 & 0.8401748932 & \bfseries 0.859618715 & 34.71409486 & 131.0021331 & 148.8079431 & 0.9993006792 & 991\\
life expectancy & 21 & 1649 & 0.2458027286 & 0.581909461 & \bfseries 0.598744696 & 6.425559049 & 68.37190909 & 161.1669742 & 0.9554683484 & 245\\
world happiness & 8 & 315 & 0.2737710177 & 0.599353052 & \bfseries 0.6529582332 & 1.757577017 & 5.242041636 & 34.33690858 & 0.969110393 & 108\\

\bottomrule
\end{tabular}
\end{small}
\caption{Empirical results overview. Table shows area under \#rules/score curve and computation time for \rulefit (rf), greedy rule boosting (grd), and optimal rule boosting (opt).
As reference, also prediction score and number of rules for \rulefit with unbounded number of rules are given.
Score is ROCAUC for classification (upper section) or $R^2$ for regression (lower section). All results are mean values over five different random 80/20 train/test splits.}
\label{tab:res_summary}
\end{table*}
In this section we present comparative results of optimal and greedy rule boosting as well as \rulefit when fitting small rule ensembles to a range of synthetic and real-world prediction problems.
Here, \textit{greedy rule boosting} refers to using the standard rule learner that, starting from the empty conjunction, chooses one proposition at a time  maximizing the increase in \eqref{eq:obj} and stops if no improvement can be achieved.
In addition to the predictive performance we also assess the learners' computational efficiency.
To that end, we use a publicly available preliminary Python implementation of the boosting algorithms and the  implementation of \rulefit available on PyPi, which is based on the highly efficient \xgboost for obtaining the input forest
(see Supplementary Materials for additional details, links, and results).

\begin{figure*}[t]
    \centering
    \begin{subfigure}[c]{0.32\textwidth}
    \centering
    \includegraphics[width=\textwidth]{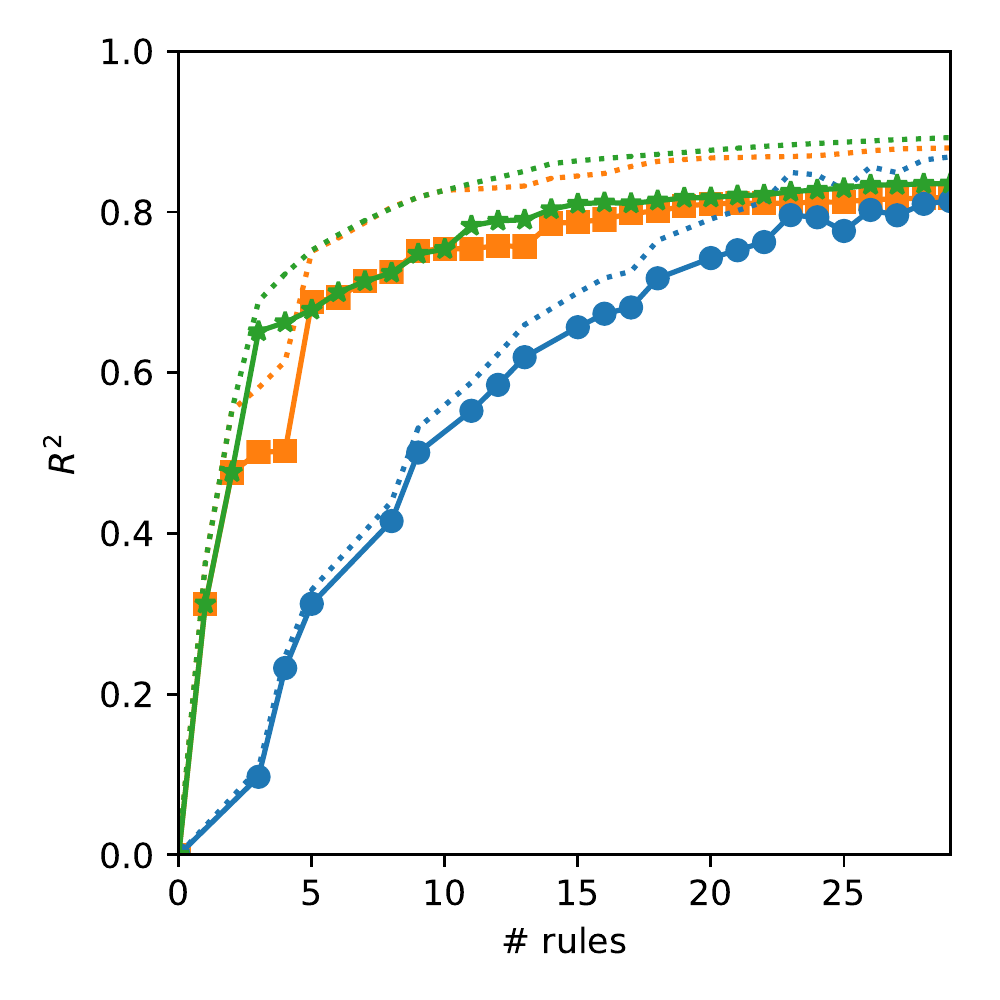}
    \caption{cars}
    \label{subfig:cars_size_perf}
    \end{subfigure}
    \hfill
    \begin{subfigure}[c]{0.32\textwidth}
    \centering
    \includegraphics[width=\textwidth]{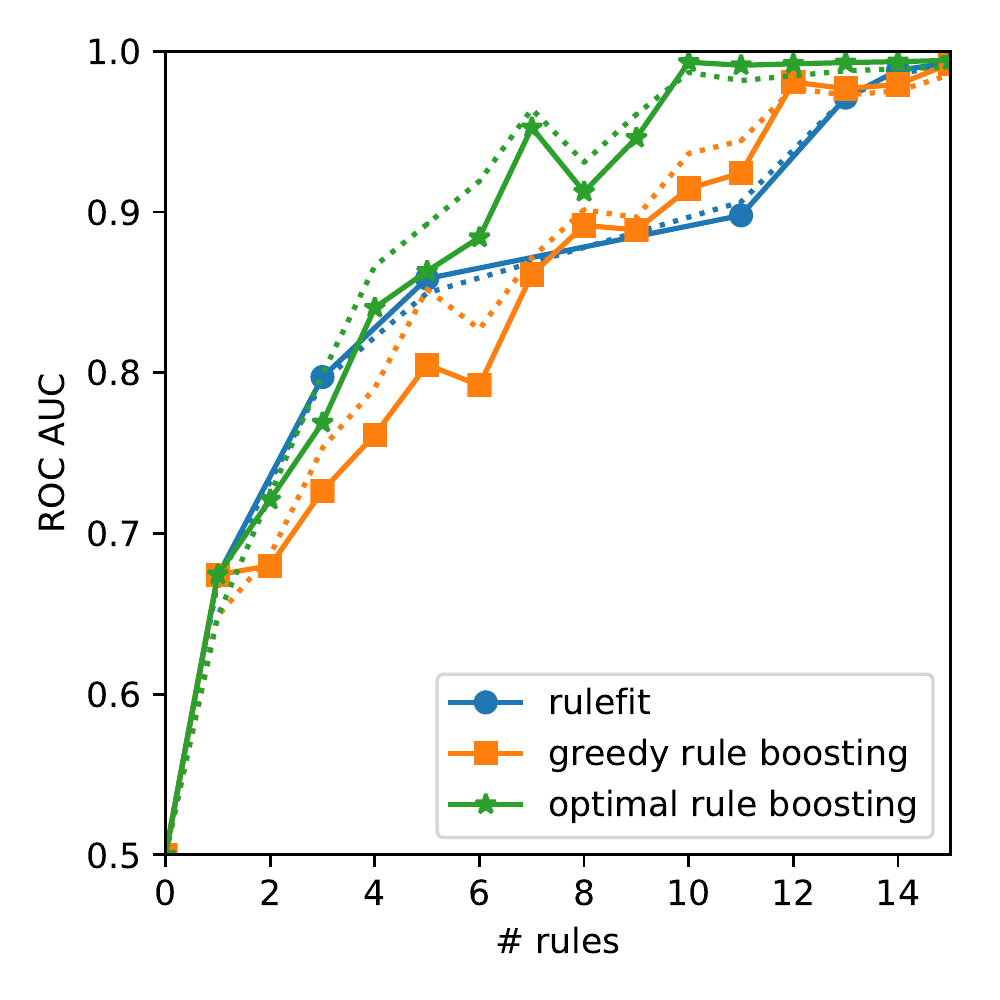}
    \caption{tic-tac-toe}
    \label{subfig:tic-tac-toe_size_per}
    \end{subfigure}
    \hfill
    \begin{subfigure}[c]{0.32\textwidth}
    \centering
    \includegraphics[width=\textwidth]{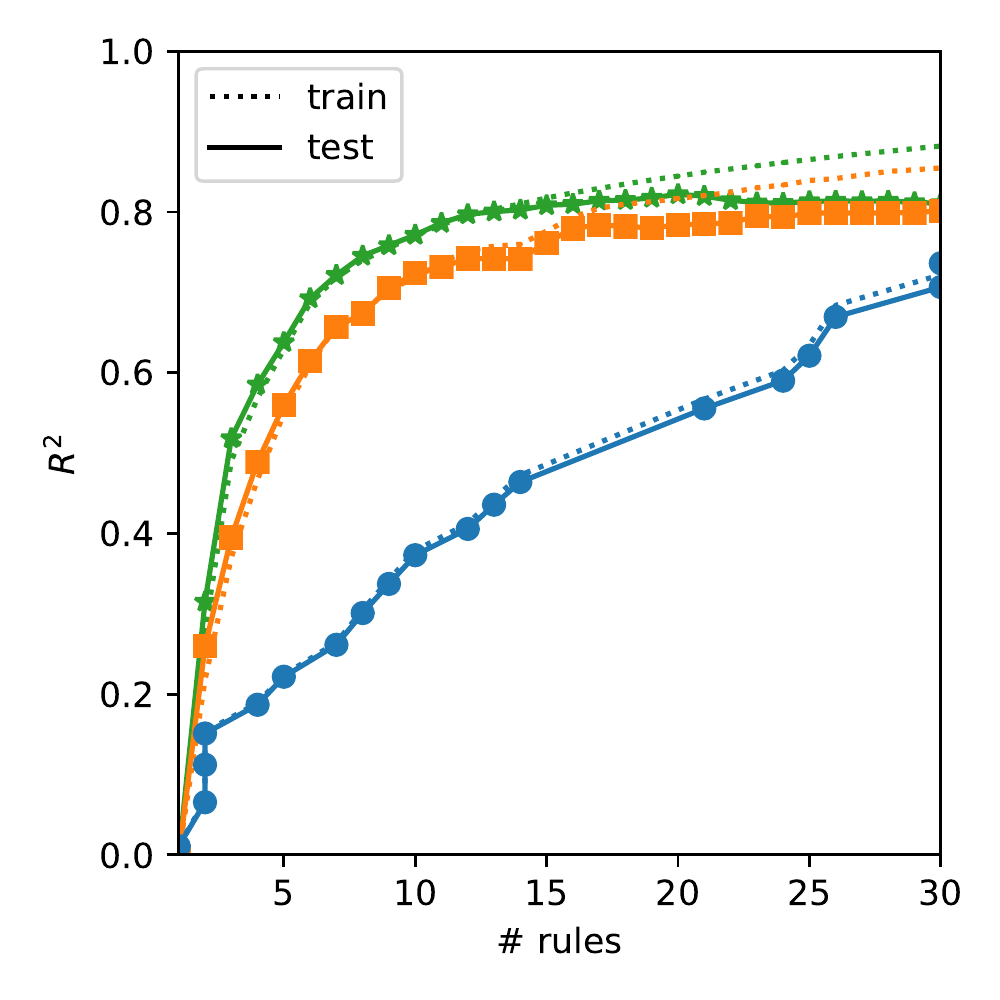}
    \caption{friedman1}
    \label{subfig:dim_vs_acc_Fried}
    \end{subfigure}
    \caption{
    Predictive performance against number of rules for selected prediction problems.
    }
    \label{fig:size_perf_curves}
\end{figure*}

\paragraph{Setting} Based on our goal of producing comprehensible models, we investigate the rule learner's performance for small ensemble sizes of up to ten rules.
While the boosting algorithms can naturally produce any desired ensemble size, for \rulefit this entails tuning the l1-regularization parameter to realize as many small ensemble sizes as possible (and average the performance for specific sizes).
Since this approach does not necessarily yield all ensemble sizes from $1$ to $10$, we interpolate performances for missing sizes by considering the area under the size/performance curve (see Fig.~\ref{fig:size_perf_curves}).

Specifically, for regression problems we measure the models' $R^2$ score, and for classification problems we measure the area under the ROC curve.
The required computation time is assessed for the greatest ensemble size (10).
For all prediction problems, the measured performance is averaged over five repetitions with different training/test splits.
The selected prediction problems contain all problems included in the \texttt{scikit-learn.datasets} module and, motivated by their interpretability, the most popular problems from Kaggle competitions.

\paragraph{Overall Results}
The overall results are summarized in Tab.~\ref{tab:res_summary} along with the basic problem characteristics.
In terms of predictive performance, we find that optimal rule boosting  consistently outperforms greedy rule boosting, which in turn outperforms the \rulefit models, for all 24 problems (\emph{rendering the hypothesis that $S_\text{opt} > S_\text{grd} > S_\text{rf}$ significant at the 0.01-level based on two one-sided sign-tests; where $S$ is the average performance over 1-10 rules of the method on a random dataset}).
While for five problems, optimal and greedy boosting are essentially en par, for 19 problems the average performance difference is more than $0.01$, and for four out of those the difference is more than $0.03$ score points.
Importantly, these results reflect average differences across ensembles sizes. For individual sizes the advantage can be much more pronounced (see below).


In terms of computation time we can observe that, while greedy boosting generally outperforms optimal boosting,  the required computation times are in the same order of magnitude (less than a factor 5 apart) for 13 of the problems.
For eight further problems, greedy is one order of magnitude faster than optimal (factor between 5 and 50), and only for three problems (\textit{IBM-HR}, \textit{boston}, \textit{mobile prices}) optimal rule boosting requires two orders of magnitude more computation time.
Importantly, for all but one problem, an optimal rule boosting ensemble is found in less than one hour of computation time on a personal computer without a highly optimized implementation.




\paragraph{Analysis}
When analyzing the predictive figures in more detail, we find that optimal and greedy ensembles generally have a lot of rules in common.
However, an individual sub-optimal rule in an ensemble can be enough to set greedy ensembles behind.
For instance for the \dataset{used cars} price regression problem, the $5$-rule boosting ensembles for the two base learners are:
\begin{small}
\begin{verbatim}
optimal base learner
1: +16192 if PS>=100 & year>=2003
2: +10596 if count<=86 & PS>=180 & year>=2009
3: - 8360 if PS in [100,180] & year in [2003,2009]
4: + 5837 if PS>=180 & year<=2003
5: + 2497 if km<=70000

greedy base learner
1: +16202 if PS>=100 & year>=2003
2: +10612 if count<=86 & PS>=180 & year>=2009
3: + 9791 if year>=2015
4: + 5790 if PS>=180 & year<=2003
5: - 8414 if PS in [100,180] & year in [2003,2009]
\end{verbatim}
\end{small}
That is, both base learners yield the same two initial rules. At position $3$, however, the complex interaction between engine power and construction year is not (yet) discovered by the greedy base learner. Instead, it is replaced by a single condition rule with much smaller predictive gain, which accounts for a substantial performance difference for ensemble sizes $3$ and $4$; until the rule is finally discovered at position $5$ (see Fig.~\ref{subfig:cars_size_perf}).


Another example is the \dataset{tic-tac-toe} classification problem, where for size $5$ we end up with the ensembles:
\begin{small}
\begin{verbatim}
optimal base learner
1: +1.187 if mid-mid==x
2: +1.717 if top-lft==x & top-mid==x & top-rgt==x
3: +1.721 if bot-lft==x & bot-mid==x & bot-rgt==x
4: +1.671 if bot-rgt==x & mid-rgt==x & top-rgt==x
5: -2.344 if top-lft==o & top-mid==o & top-rgt==o

greedy base learner
1: +1.187 if mid-mid==x
2: +1.211 if bot-rgt==x & mid-mid==b
3: -2.671 if top-lft==o & top-mid==o & top-rgt==o
4: +1.487 if bot-lft==x & bot-mid==x & bot-rgt==x
5: +1.685 if top-lft==x & top-mid==x & top-rgt==x
\end{verbatim}
\end{small}
That is, after the initial ``statistical'' rule, optimal rule boosting proceeds with adding exact ``symbolic'' win conditions to its ensembles.
Greedy rule boosting creates a very similar ensemble but inserts a largely ineffective statistical rule at position $2$, which again leads to a size/performance disadvantage (see Fig.~\ref{subfig:tic-tac-toe_size_per}).
   



\section{Conclusions}
Our analysis of the second order gradient boosting objective yields a computationally efficient optimal base learner for rule boosting.  
As our results indicate, this base learner consistently outperforms a conventional greedy rule learner in terms of its predictive performance.
Besides the previous lack of an efficient search algorithm, the advantage of an optimal base learner might have been overlooked thus far, because it tends to vanish for large ensemble sizes.
However, as demonstrated by the noisy parity example of Fig.~\ref{fig:parity_example}, 
some problems with higher-order interactions are learned accurately by optimal rule boosting with relatively few rules but are intractable by greedy rule boosting even for very large ensemble sizes.

Thus, the presented approach adds a tool of general interest to the boosting framework
that should in particular be considered when seeking comprehensible, i.e., small, models:
Not only are the predictive gains most pronounced for a small number of rules. In this regime, also the required computational overhead can be tolerated more easily.
One direction to push the comprehensibility/accuracy trade-off even further, is to completely replace the stage-wise fitting paradigm and consider ``global optimization'' approaches.
An efficient algorithm for this has been proposed for rule \emph{lists} and the specific case of accuracy-based classification~\citep{angelino2017learning}.
Lifting these ideas to additive rule ensembles, ideally of the same universality as the gradient boosting framework, is a natural next goal.


\begin{footnotesize}
\setlength{\bibsep}{0pt}
\setstretch{0.95}
\bibliographystyle{abbrvnat}
\bibliography{biblio}
\end{footnotesize}

\appendix

\section{Technical Results}

\begin{figure}
    \centering
    \includegraphics[width=1.1\columnwidth]{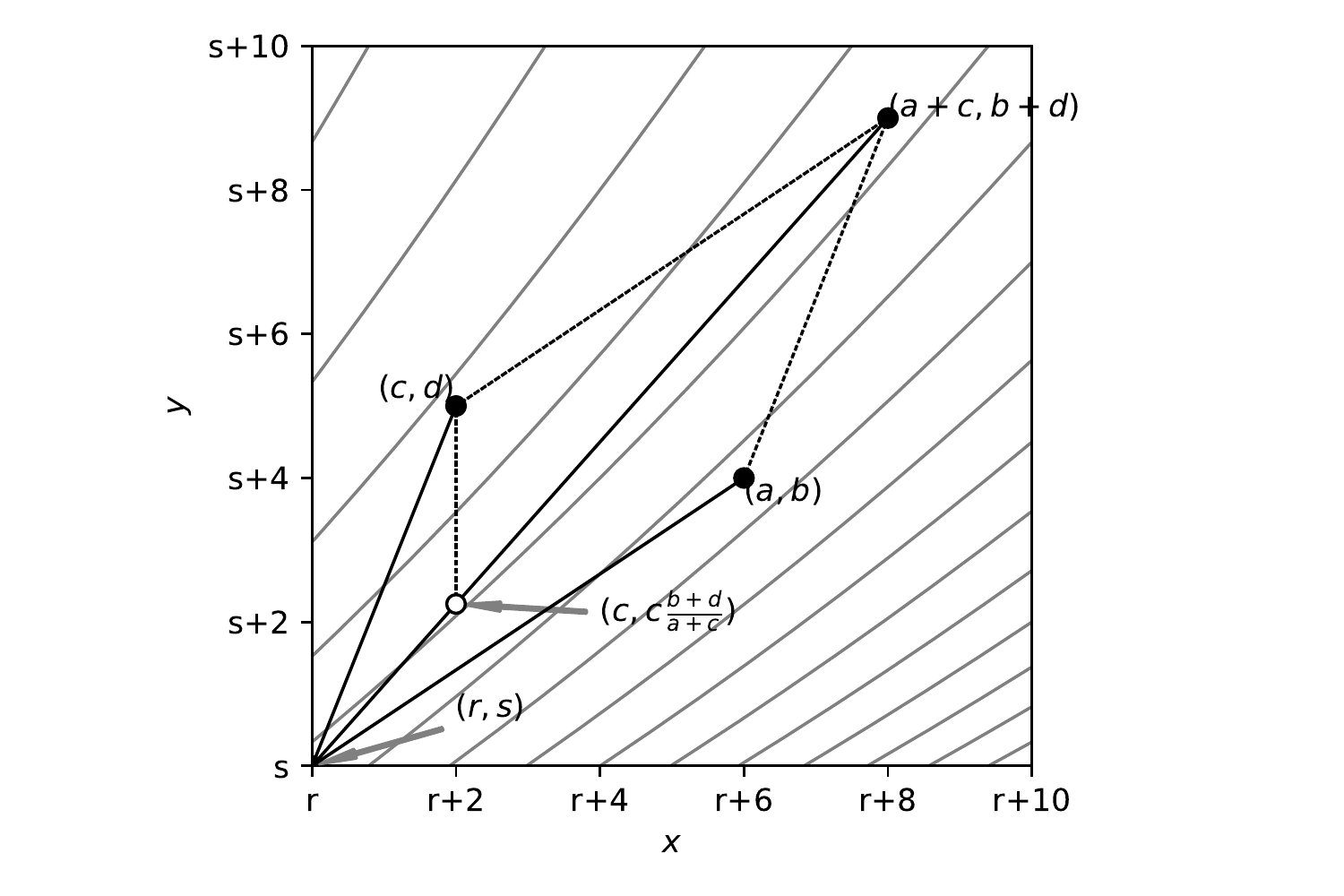}
    \caption{Illustration of the proof of Lemma~\ref{lm:curvature}.}
    \label{fig:proof}
\end{figure}

\begin{lemma}
Let $r$, $a$, and $c$ be positive real numbers, and $s$, $b$, and $d$ be strictly positive numbers such that $a/b > c/d$. It holds that
\begin{equation}
    \frac{(r+c)^2}{s + d} \leq \max \left\{ \frac{r^2}{s}, \frac{(r+a+c)^2}{s + b + d} \right\} \enspace .
\end{equation}
\label{lm:curvature}
\end{lemma}
\begin{proof}
Setting $x = a + c$, and $y = b + d$, we define for $\alpha \in [0,1]$ the function
\[
z(\alpha) = \frac{(r + \alpha x)^2}{s + \alpha y} \enspace .
\]
See Fig.~\ref{fig:proof} for an illustration.
With this definition we have $z(0)=r^2/s$, $z(1)=(r+a+c)^2 / (s+b+d)$ and
\begin{align*}
z\left( \frac{c}{a+c} \right) & = \frac{(r+c)^2}{ s + c(b+d)/(a+c)} \\
& \geq \frac{(r+c)^2}{s+d} 
\enspace ,
\end{align*}
where the inequality holds because of our premise that $a/b \geq c / d$, or equivalently that $b/a \leq d/c$, which implies that
\[
s + d = s + c \frac{d}{c} \geq s + c\frac{b+d}{a+c} \enspace .
\]

Furthermore, $z(\alpha)$ is convex, as we can see by finding its second derivative
\[
z''(\alpha) = \frac{2(ry - sx)^2}{(s + \alpha y)^3} \enspace ,
\]
which is positive for $\alpha \in [0,1]$ based on the positivity of $s$, $b$, and $d$. The convexity of $z$ implies that its maximum is attained at the boundary, i.e., either for $\alpha = 0$ or $\alpha=1$. Thus, 
\begin{align*}
    \frac{(r+c)^2}{s+d} & \leq z\left( c/(a+c) \right) \\
    & \leq \max \{z(0), z(1)\}\\
    & = \max \{r^2/s, (r+a+c)^2 / (s+b+d)\}
\end{align*}
as required.
\end{proof}

\section{Details of Experimental Setup}
The code of all experiments as well as exact versions of datasets and software dependencies can be retrieved from \url{https://github.com/SimonTeshuva/interpretable_Rule_Ensemble}.

\paragraph{Dataset selection} The empirical study targeted interpretable datasets of small to medium size. For that, all datasets from \texttt{sklearn.datasets} have been included that are loaded by a function with name starting \texttt{load$\_$}. That is, the large datasets that are ``fetched'' online have been excluded. Additionally, all Kaggle datasets with more than 300 votes (as of December 2019) have been retrieved filtered by the following criteria: 
\begin{enumerate}
    \item Competition has a clearly defined regression or classification problem given in a csv file.
    \item File is between 50KB and 1MB.
    \item For similar datasets only the higher ranked one was selected.
\end{enumerate}

\paragraph{Rule Learner Configuration}
For both greedy and optimal rule boosting, the regularization parameter of the base learner was optimized from the following options:
\begin{equation*}
\lambda \in \{0.0001, 0.001, 0.01 , 0.1 , 0.2, 0.5, 1, 2, 5, 10, 20, 50\} \enspace ,
\end{equation*}
where for the optimal base learner some small options were skipped whenever the required computation time exceeded 2h (see additional experiments below for a discussion of the effect of $\lambda$ on the computation time).
For \rulefit, first boundary regularization values were found that create $1$ and approximately $11$ rules, respectively. Then, the algorithm was run with $50$ regularization values that linearly interpolate between these boundary values, resulting in the reported average performance.

\paragraph{Computing Environment} All experiments were carried out on a personal laptop computer equipped with an Intel i5-6200U (4) @ 2.800GHz CPU, an integrated Intel Skylake GT2 GPU
and 8GB RAM.
The system was running Ubuntu 18.04.4 LTS x86\_64 and Python: 3.6.9 .

\section{Additional Experiments}
\begin{figure}[htb]
    \centering
    \includegraphics[width=\columnwidth]{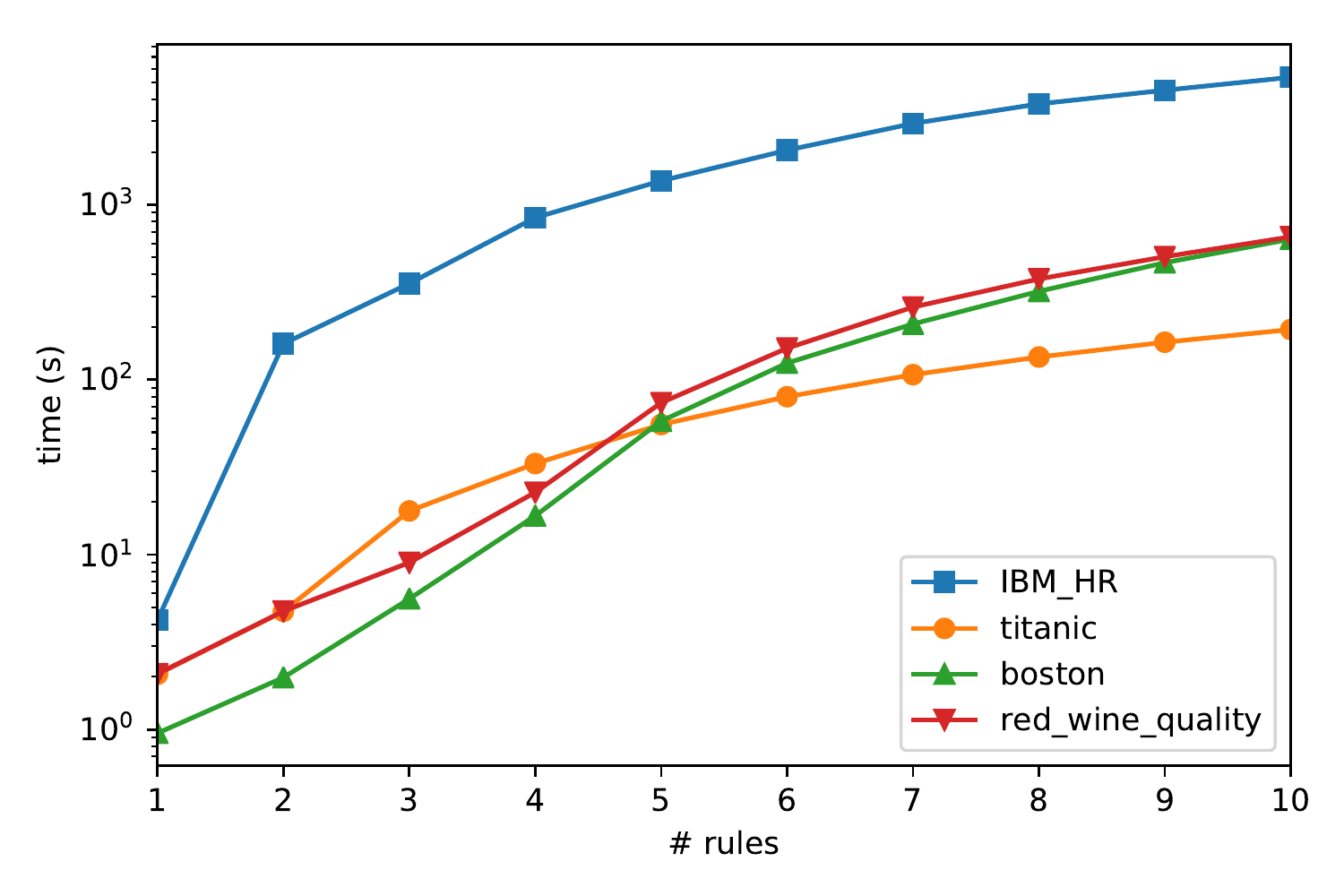}
    \caption{Computation time of optimal rule boosting versus number of rules.}
    \label{fig:time_vs_rules}
\end{figure}
The experimental results presented in the main text are mainly focused on predictive performance and only include summary computation times for the completely solving the considered prediction tasks.
Beyond this, it is of practical interest to further assess the factors with which one can influence the computation time of optimal rule boosting.

\paragraph{Scaling in number of rules} A first question is how the computation time scales with the number of rules.
Naively one could assume a simple linear dependence. However, we typically find a super-linear behavior where the required computation time increases very fast for the first couple of rules and then starts to level off for larger ensemble sizes (see Fig.~\ref{fig:time_vs_rules}).
This phenomenon can be explained by the coverage of the discovered rules: In the first couple of iterations there are typically still strong general rules that are identified early in the search process such that a lot of low coverage rules do not have to be explored due to the optimistic estimator pruning.
Once these patterns are exhausted, the search has to explore more specific rules, which suffers from the corresponding combinatorial growth of the relevant part of the search space.

\begin{figure}[htb]
    \centering
    \includegraphics[width=\columnwidth]{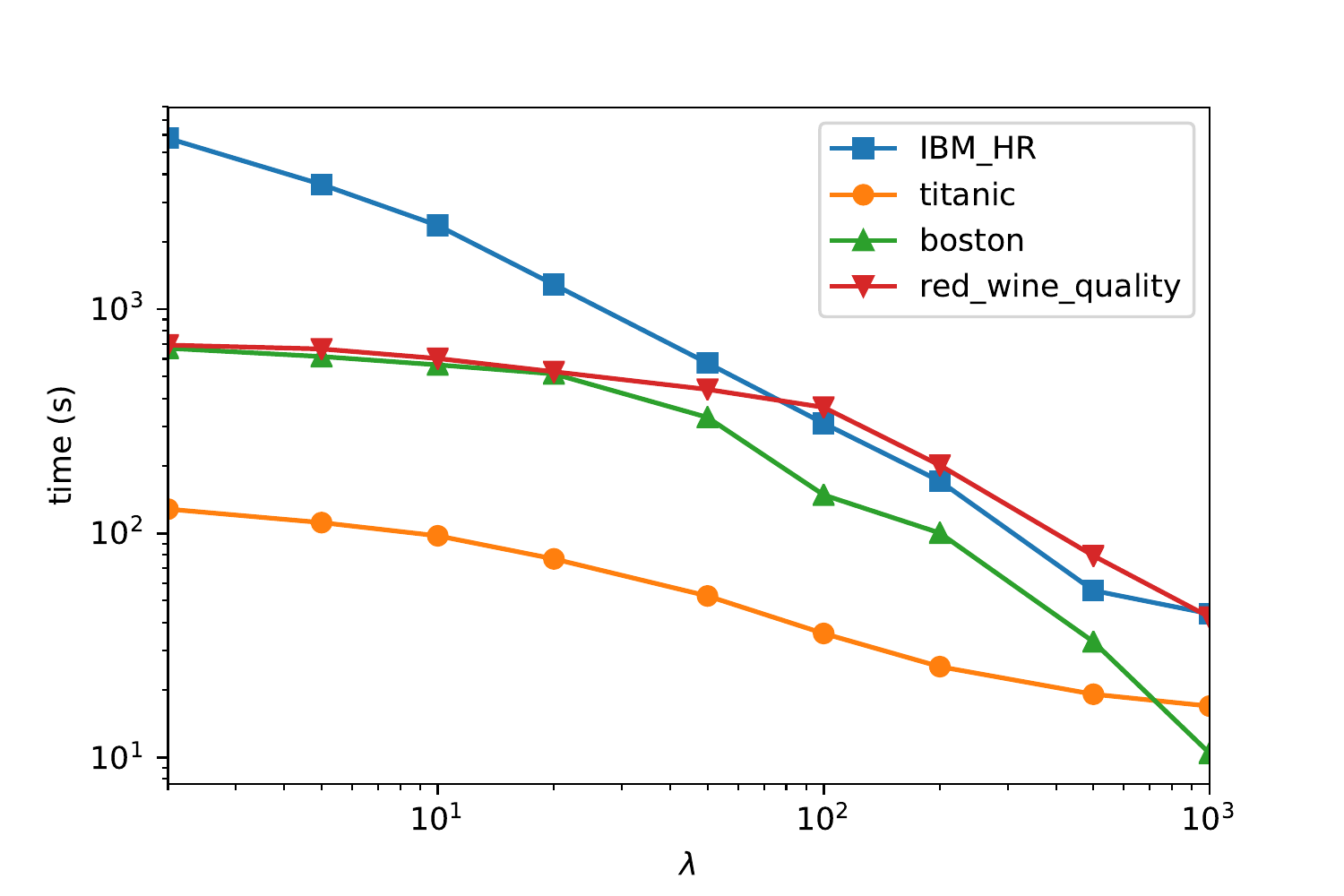}
    \caption{Computation time of optimal rule boosting (ensemble size 10) versus regularization parameter $\lambda$.}
    \label{fig:time_vs_lambda}
\end{figure}
\paragraph{Scaling in regularization parameter} This observation suggest to use the regularization parameter $\lambda$ as a principled means to reduce the required computation time (in addition to relaxing the exact optimization to an  approximation with multiplicative guarantee, as briefly discussed in the main text).
Analyzing the objective function, we find that the gain for covering additional data point (with the same sign as the overall gradient sum) increases with $\lambda$.
Consequently, greater $\lambda$ values result in optimal rules with greater coverage, which again allows additional pruning and, hence, lower computation time.
Fig.~\ref{fig:time_vs_lambda} shows this effect for the same datasets as considered above.

\paragraph{Effect of pruning techniques} While the above two insights are perhaps the most important from a user perspective, for guiding further algorithmic development it is also instructive to investigate
the relative importance of the different pruning mechanism employed in Algorithm 1.
We can distinguish between the two mechanisms of bounding the objective function (enabled by Thm.~3.1) and avoiding the redundant enumeration of equivalent queries (through the core query construct of LCM).
Moreover, it is instructive to differentiate the immediate application of these mechanism as in standard branch-and-bound and the propagated application in OPUS, which in the case of equivalence pruning is enabled by Thm.~3.2.

Tab.~\ref{tab:pruning_rules} contains the number of activations of each of these mechanisms for eight exemplary datasets.
We can generally confirm previous studies that found that typically bounding and non-redundancy contribute to pruning in the same order of magnitude (although variations are possible where either mechanism is much more important than the other).
As an interesting novel observation, we can see that propagating equivalence-related pruning information through the critical index construction can be equally effective as propagation of the objective bound (note that the table gives a conservative estimate, as additional non-recursive pruning enabled by (3.8) is not included in these numbers).
This observation underlines the contribution of Thm.~3.2 to enable an overall efficient rule optimization.
\begin{table}[htb]
    \centering
    \begin{small}
    \begin{tabular}{lrrrr}
    \toprule
    {\bf dataset} & \multicolumn{2}{c}{\bf immediate} & \multicolumn{2}{c}{\bf propagated}\\
     & bnd & equiv & bnd & equiv\\
    \midrule
boston	& 1280132 & 941252 & 146216 &	405358\\
breast & 151420 & 88881 & 18902 & 54427\\
friedman1 & 227373 & 4473 & 2981 & 0\\
iris & 15881 & 26208 & 2255 & 3310\\
red wine qu. & 2881173 & 149112 & 54389 & 114079\\
tic-tac-toe	& 4303 & 7772 & 2179 & 0\\
titanic & 106374 & 911809 & 1969 & 27241\\
used cars & 1347 & 990 & 183 & 1\\
wrl. happi. & 148327 & 17080 & 26420 & 5904\\
    \bottomrule
    \end{tabular}
    \end{small}
    \caption{Number of pruning rule activations: bounding the objective function (bnd) and avoiding redundant generation of equivalent queries (equiv), separated by immediate application and through OPUS propagation of pruning information (propagated).}
    \label{tab:pruning_rules}
\end{table}

\end{document}